\documentclass[11pt,letterpaper]{article}
\usepackage[letterpaper]{geometry}
\usepackage{acl2012}
\usepackage{times}
\usepackage{latexsym}
\usepackage{amsmath}
\usepackage{amssymb}
\usepackage{mathtools}
\usepackage{multirow}
\usepackage{todonotes}
\usepackage{amssymb}
\usepackage{bbm}
\usepackage{url}
\usepackage{graphics}
\usepackage{comment}
\usepackage{subcaption}
\usepackage{subfig}
\usepackage{graphicx}
\usepackage{algorithmicx}
\usepackage[ruled]{algorithm}
\usepackage[noend]{algpseudocode}
\usepackage{multirow}
\usepackage[titletoc,toc,title]{appendix}
\usepackage{textcomp}

\makeatletter

\newcommand{\@BIBLABEL}{\@emptybiblabel}
\newcommand{\@emptybiblabel}[1]{}
\makeatother

\title{Cross-Lingual Syntactic Transfer with Limited Resources}

 \author{Mohammad Sadegh Rasooli \and Michael Collins\thanks{~~On leave at Google Inc. New York.}\\
   Department of Computer Science,
   Columbia University \\
   New York, NY 10027, USA \\
  {\tt \{rasooli,mcollins\}@cs.columbia.edu}  \\}

\date{}

\begin{document}
\maketitle

\begin{abstract}
We describe a simple but effective method for cross-lingual syntactic
transfer of dependency parsers, in the scenario where a large amount
of translation data is not available. The method makes use of three
steps: 1) a method for deriving cross-lingual word clusters, which can
then be used in a multilingual parser; 2) a method for transferring
lexical information from a target language to source language
treebanks; 3) a method for integrating these steps with the
density-driven annotation projection method of
\newcite{rasooli-collins:2015:EMNLP}. Experiments show improvements
over the state-of-the-art in several languages used in previous work, 
%\cite{rasooli-collins:2015:EMNLP,yuanregina15,ammar2016one}, 
in a setting where the only source of translation data is
the Bible, a considerably smaller corpus than the Europarl corpus used
in previous work. Results using the Europarl corpus as a source of
translation data show additional improvements over the results of
\newcite{rasooli-collins:2015:EMNLP}. We conclude with results on 38
datasets from the Universal Dependencies corpora.

% 13 datasets (10 languages) have unlabeled attachment accuracies of 80\%
%or higher; the average unlabeled accuracy on the 38 datasets is 74.8\%.

\end{abstract}

\section{Introduction} \label{intro}

Creating manually-annotated syntactic treebanks is an expensive and
time consuming task.  Recently there has been a great deal of interest
in cross-lingual syntactic transfer, where a parsing model is trained
for some language of interest, using only treebanks in other
languages.  There is a clear motivation for this in building parsing
models for languages for which treebank data is unavailable. Methods
for syntactic transfer include annotation projection methods
\cite{hwa2005bootstrapping,ganchev-gillenwater-taskar:2009:ACLIJCNLP,mcdonald-petrov-hall:2011:EMNLP,ma-xia:2014:P14-1,rasooli-collins:2015:EMNLP,lacroix-EtAl:2016:N16-1,agic2016multilingual},
learning of delexicalized models on universal treebanks \cite{zeman2008cross,mcdonald-petrov-hall:2011:EMNLP,tackstrom2013target,rosa-zabokrtsky:2015:ACL-IJCNLP}, treebank translation \cite{tiedemann-agic-nivre:2014:W14-16,tiedemann2015improving,tiedemann2016synthetic} and
methods that leverage cross-lingual representations of word clusters, embeddings or dictionaries \cite{tackstrom2012cross,durrett-pauls-klein:2012:EMNLP-CoNLL,duong-EtAl:2015:CoNLL,yuanregina15,xiao-guo:2015:CoNLL,guo-EtAl:2015:ACL-IJCNLP2,guo2016representation,ammar2016one}.

This paper considers the problem of cross-lingual syntactic transfer
with limited resources of monolingual and translation
data. Specifically, we use the Bible corpus of \newcite{christodouloupoulos2014massively} as a
source of translation data, and Wikipedia as a source of monolingual
data. 
We deliberately
limit ourselves to the use of Bible translation data because it is
available for a very broad set of languages: the data from
\newcite{christodouloupoulos2014massively} includes data from 100 languages. 
The Bible data 
contains a much smaller set of
sentences (around 24,000) than other translation corpora, for example
Europarl \cite{koehn2005europarl}, which has around 2 million sentences per language pair. 
This makes it a considerably more challenging corpus to work with.
Similarly, our choice of Wikipedia as the source of monolingual
data is motivated by the availability of Wikipedia data in a very
broad set of languages.

We introduce a set of simple but effective methods for
syntactic transfer, as follows:

\begin{itemize}

\item We describe a method for deriving cross-lingual clusters, where
  words from different languages with a similar syntactic or semantic
  role are grouped in the same cluster. These clusters can then be
used as features in a shift-reduce dependency parser.

\item We describe a method for transfer of lexical information from
the target language into source language treebanks, using word-to-word
translation dictionaries derived from parallel corpora. Lexical
features from the target language can then be integrated in parsing.

\item We describe a method that integrates the above two approaches
  with the density-driven approach to annotation projection
  described in \cite{rasooli-collins:2015:EMNLP}.

\end{itemize}

Experiments show that our model outperforms previous work on a set of
European languages from the Google universal treebank
\cite{mcdonald-EtAl:2013:Short}: we achieve 80.9\% average unlabeled
attachment score (UAS) on these languages; in comparison the work of
\newcite{yuanregina15}, \newcite{guo2016representation} and
\newcite{ammar2016one} have UAS of 75.4\%, 76.3\% and 77.8\%
respectively. All of these previous works make use of the much larger
Europarl \cite{koehn2005europarl} corpus to derive lexical
representations.  When using Europarl data instead of the Bible, our
approach gives 83.9\% accuracy, a 1.7\% absolute improvements over
\cite{rasooli-collins:2015:EMNLP}. Finally, we conduct experiments on
38 datasets (26 languages) in the universal dependencies v1.3 \cite{11234/1-1699}
corpus. Our method has an average unlabeled dependency accuracy of 74.8\% for
these languages, more than 6\% higher than the method
of \newcite{rasooli-collins:2015:EMNLP}. 13 datasets (10 languages) have accuracies higher than 80.0\%.\footnote{~The parser code is available at \url{https://github.com/rasoolims/YaraParser/tree/transfer}.}

\section{Background}
\label{background_section}

\newcommand{\gen}{{\cal Y}}
\newcommand{\reals}{\mathbb{R}}
\newcommand{\tb}{{\cal T}}
\newcommand{\lang}{{\cal L}}
\newcommand{\mono}{{\cal D}}
\newcommand{\bible}{{\cal B}}
\newcommand{\nullt}{{\tt NULL}}
\newcommand{\embed}{v}

This section gives a description of the underlying parsing models used
in our experiments, the data sets used, and a baseline approach based
on delexicalized parsing models.

\subsection{The Parsing Model}

We assume that the parsing model is a discriminative linear model,
where given a sentence $x$, and a set of candidate parses $\gen(x)$, 
the output from the model is
\[
y^*(x) = \arg\max_{y \in \gen(x)} \theta \cdot \phi(x, y)
\]
where $\theta \in \reals^d$ is a parameter vector, and $\phi(x, y)$ is
a feature vector for the pair $(x, y)$.  In our experiments we use the
shift-reduce dependency parser of \newcite{rasooli2015yara}, which is an extension of the approach
in \cite{zhang-nivre:2011:ACL-HLT2011}. The parser is trained using
the averaged structured perceptron \cite{collins:2002:EMNLP02}.

We assume that the feature vector $\phi(x, y)$ is the concatenation of
three feature vectors:

\begin{itemize}

\item $\phi^{(p)}(x, y)$ is an unlexicalized set of features. Each such feature may
  depend on the part-of-speech (POS) tag of words in the sentence, but
does not depend on the identity of individual words in the sentence.

\item $\phi^{(c)}(x, y)$ is a set of cluster features. These features require
  access to some dictionary that maps each word in
  the sentence to an underlying cluster identity. Clusters may for
  example be learned using the Brown clustering algorithm
  \cite{brown1992class}. The features may make use of cluster
  identities in combination with POS tags.

\item $\phi^{(l)}(x, y)$ is a set of lexicalized features. Each
such feature may depend directly on word identities in the sentence.
These features may also depend on part-of-speech tags or
  cluster information, in conjunction with lexical information.

\end{itemize}

Appendix~\ref{apen_feat_sec} has a full description of the features used in our experiments.

\subsection{Data Assumptions}

Throughout this paper we will assume that we have $m$ source languages
$\lang_1 \ldots \lang_m$, and a single target language $\lang_{m+1}$.
We assume the following data sources:

\paragraph{Source language treebanks.} We have a treebank $\tb_i$ for each
  language $i \in \{1 \ldots m\}$.

\paragraph{Part-of-speech (POS) data.} We have hand-annotated POS data for
  all languages $\lang_1 \ldots \lang_{m+1}$. We assume that the data uses a
    universal POS set that is common across all languages.

\paragraph{Monolingual data.} We have monolingual raw data for each of the
  $(m+1)$ languages. We use $\mono_i$ to refer to the monolingual data for the $i$'th language.

\paragraph{Translation data.} We have translation data for all language
  pairs.  We use $\bible_{i,j}$ to refer to translation data for the
  language pair $(i, j)$ where $i, j \in \{ 1 \ldots (m+1)\}$ and
$i \neq j$.

In our main experiments we use the Google universal treebank
\cite{mcdonald-EtAl:2013:Short} as our source language
treebanks\footnote{We also train our best performing model on the
  newly released universal treebank v1.3 \cite{11234/1-1699}. See
  \S\ref{sec_univ_res} for more details.} (this treebank provides
universal dependency relations and POS tags), Wikipedia data as our
monolingual data, and the Bible data from
\newcite{christodouloupoulos2014massively} as the source of our
translation data. In additional experiments we use the Europarl
corpus as a source of translation data, in order to measure the impact
of using the smaller Bible corpus.

% \begin{description}

% \item[Source language treebanks and POS data] We use the Google universal treebank as our source treebanks (t
% in all our experiments. This treebank provides universal dependency relations and  
% POS tags), .

% %\item[Part-of-speech (POS) data] We again use the Google universal treebank.

% \item[Monolingual and translation data] We make use of Wikipedia data as our monolingual data, and for our
%   experiments. We use the Bible as the source of all our 
%   translation data. 

% \end{description}

% We make use of Bible translation data because it is available for a very broad set
% of languages; we are interested in developing algorithms that are
% applicable to a large number of languages, including languages where
% translation data other than the Bible is unavailable.  The Bible
% contains a much smaller set of sentences (around 24,000)
% than other translation corpora, for example Europarl, which has around
% 2 million sentences per language pair. The Bible is therefore a significantly more
% challenging data set to work with. Similarly, our choice of Wikipedia as the source of monolingual data
% is motivated by the availability of Wikipedia data in a very broad set
% of languages.

\subsection{A Baseline Approach: Delexicalized Parsers with Self-Training}

Given the data assumption of a universal POS set, the feature vectors
$\phi^{(p)}(x, y)$ can be shared across languages. A simple approach
is then to simply train a delexicalized parser using treebanks $\tb_1
\ldots \tb_m$, using the representation $\phi(x, y) = \phi^{(p)}(x,
y)$ (see \cite{mcdonald-EtAl:2013:Short,tackstrom2013target}).

Our baseline approach makes use of a delexicalized parser, with two
refinements: 

\paragraph{WALS properties.}
We use the six properties from the world atlas of language structures
(WALS) \cite{dryer2005world} to select a subset of closely related
languages for each target language. These properties are shown in
Table \ref{wals_tab}. The model for a target language is trained on
treebank data from languages where at least 4 out of 6 WALS properties
are common between the source and target language.\footnote{There was
no effort to optimize this choice; future work may consider more
sophisticated sharing schemes.}
 This gives a
slightly stronger baseline: our experiments showed an improvement in
average labeled dependency accuracy for the languages from 62.52\% to
63.18\%. Table~\ref{tab_source} shows the set of source languages used
for each target language; these source languages are used for all
experiments in the paper.
  
  \begin{table}
    \begin{tabular}{l|l}
    \hline
        Feature & Description \\ \hline
82A & Order of subject and verb \\
83A & Order of object and verb \\
85A & Order of adposition and noun phrase \\
86A & Order of genitive and noun \\
87A & Order of adjective and noun \\ 
88A & Order of demonstrative and noun \\ \hline
    \end{tabular}
    \caption{\label{wals_tab} The six properties from the world atlas of language structures (WALS) \protect \cite{dryer2005world} used to select the source languages for each target language in our experiments.}
  \end{table}

\paragraph{Self-training.}
We use self-training \cite{McClosky:2006:ESP:1220835.1220855} to
further improve parsing performance. Specifically, we first train a
delexicalized model on treebanks $\tb_1 \ldots \tb_m$; then use the
resulting model to parse a dataset $\tb_{m+1}$ that includes
target-language sentences which have POS tags but do not have
dependency structures. We finally use the automatically parsed data
$\tb'_{m+1}$ as the treebank data and retrain the model; this last
model is trained using all features (unlexicalized, clusters, and
lexicalized). Self-training in this way gives an improvement in
labeled accuracy from 63.18\% to 63.91\%.

\begin{table}[t!]
    \centering
    \begin{tabular}{l l}
  \hline
    Target & Sources \\
    \hline
    en & de, fr, pt, sv \\
    de  & en, fr, pt \\
    es & fr, it, pt \\
    fr & en, de, es, it, pt, sv \\
    it & es, fr, pt \\
    pt & en, de, es, fr, it, sv \\
    sv & en, fr, pt \\ \hline
    \end{tabular}
    \caption{The selected source languages for each target language in the Google universal treebank v2 \protect\cite{mcdonald-EtAl:2013:Short}. A language is chosen as a source language if it has at least $4$ out of $6$ WALS properties in common with the target language.}
    \label{tab_source}
\end{table}

%Column 1 of table~\ref{tab_results1} shows the final dependency parsing accuracy (UAS and LAS) of this approach.

\subsection{Translation Dictionaries}

Our only use of the translation data $\bible_{i,j}$ for $i, j \in \{1
\ldots (m+1)\}$ is to construct a translation dictionary $t(w, i, j)$. Here
$i$ and $j$ are two
languages, $w$ is a word in language $\lang_i$, and the output $w' =
t(w, i, j)$ is a word in language $\lang_j$  corresponding to the most
frequent translation of $w$ into this language. 

We define the function $t(w, i, j)$ as follows. We first run the
GIZA++ alignment process \cite{och2000giza} on the data
$\bible_{i,j}$. We then keep intersected alignments between sentences
in the two languages. Finally, for each word $w$ in $\lang_i$, we
define $w' = t(w, i, j)$ to be the target language word most
frequently aligned to $w$ in the aligned data. If a word $w$ is never
seen aligned to a target language word $w'$, we define $t(w, i, j) =
\nullt$. 
%Because the Bible is very small, many words have the $\nullt$
%translation. To address this problem, we design a vocabulary expansion
%approach (see \S\ref{expansion_section}) to increase the coverage of
%the translation function.

% Future work may consider the use of probabilistic lexicons, or
% hand-crafted lexicons, as an alternative to the method described above.

\makeatletter
\newcommand{\subalign}[1]{%
  \vcenter{%
    \Let@ \restore@math@cr \default@tag
    \baselineskip\fontdimen10 \scriptfont\tw@
    \advance\baselineskip\fontdimen12 \scriptfont\tw@
    \lineskip\thr@@\fontdimen8 \scriptfont\thr@@
    \lineskiplimit\lineskip
    \ialign{\hfil$\m@th\scriptstyle##$&$\m@th\scriptstyle{}##$\crcr
      #1\crcr
    }%
  }
}
\makeatother

\section{Our Approach}

We now describe an approach that gives significant improvements over
the baseline. \S\ref{sec:clusters} describes a method for deriving cross-lingual
clusters, allowing us to add cluster features $\phi^{(c)}(x, y)$ to
the model.
 \S\ref{expansion_section} describes a method for adding lexical features
$\phi^{(l)}(x, y)$ to the model.
 \S\ref{density} describes a method for integrating the approach
with the density-driven approach of \newcite{rasooli-collins:2015:EMNLP}.
 Finally, \S\ref{sec_experiments} describes experiments. We show that each of the
above steps leads to improvements in accuracy.

\subsection{Learning Cross-Lingual Clusters}

%{\bf TODO: add section giving a justification for the clustering algorithm}

\label{sec:clusters}
\input{code_switch_pseudo}

We now describe a method for learning cross-lingual clusters.
This follows previous work on cross-lingual clustering algorithms \cite{tackstrom2012cross}.
A {\em clustering} is a function $C(w)$ that maps each
word $w$ in a vocabulary to a cluster $C(w) \in \{ 1 \ldots K\}$,
where $K$ is the number of clusters. A {\em hierarchical clustering}
is a function $C(w, l)$ that maps a word $w$ together with an integer
$l$ to a cluster at level $l$ in the hierarchy. As one example, the
Brown clustering algorithm \cite{brown1992class} gives a hierarchical clustering. The level
$l$ allows cluster features at different levels of granularity.

A {\em cross-lingual} hierarchical clustering is a function $C(w, l)$
where the clusters are shared across the $(m+1)$ languages of interest:
that is, the word $w$ can be from any of the $(m+1)$ languages.
Ideally, a cross-lingual clustering should put words across different
languages which have a similar syntactic and/or semantic role in the
same cluster.  There is a clear motivation for cross-lingual
clustering in the parsing context. We can use the cluster-based
features $\phi^{(c)}(x, y)$ on the source language treebanks $\tb_1
\ldots \tb_m$, and these features will now generalize beyond these
treebanks to the target language $\lang_{m+1}$.

We learn a cross-lingual clustering by leveraging the monolingual data
sets $\mono_1 \ldots \mono_{m+1}$, together with the translation
dictionaries $t(w, i, j)$ learned from the translation data.
Figure~\ref{cs_alg2} shows the algorithm that learns
a cross-lingual clustering. The algorithm first prepares a
multilingual corpus, as follows: for each sentence $s$ in the
monolingual data $\mono_i$, for each word in $s$, with probability
$\alpha$ we replace the word with its translation into some randomly
chosen language. Once this data is created, we can easily obtain a
cross-lingual clustering. Figure~\ref{cs_alg2} shows the complete
algorithm. The intuition behind this method is that by creating
the cross-lingual data in this way, we bias the clustering
algorithm towards putting words that are translations of each
other in the same cluster.

\subsection{Treebank Lexicalization}\label{expansion_section}

We now describe how to introduce lexical representations
$\phi^{(l)}(x, y)$ to the model. Our approach is simple: we take the
treebank data $\tb_1 \ldots \tb_m$ for the $m$ source languages,
together with the translation lexicons $t(w, i, m+1)$. For any word
$w$ in the source treebank data, we can look up its translation $t(w,
i, m+1)$ in the lexicon, and add this translated form to the
underlying sentence. Features can now consider lexical identities
derived in this way. In many cases the resulting translation will be
the {\tt NULL} word, leading to the absence of lexical
features. However, the representations $\phi^{(p)}(x, y)$ and
$\phi^{(c)}(x, y)$ still apply in this case, so the model is robust to
some words having a {\tt NULL} translation.

\subsection{Integration with the Density-Driven Projection Method of \protect\newcite{rasooli-collins:2015:EMNLP}}
\label{density}

In this section we describe a method for integrating our approach with
the cross-lingual transfer method of \newcite{rasooli-collins:2015:EMNLP}, 
which makes use of density-driven projections.

In annotation projection methods \cite{hwa2005bootstrapping,mcdonald-petrov-hall:2011:EMNLP}, it is assumed that we have
translation data $\bible_{i,j}$ for a source and target language, and
that we have a dependency parser in the source language $\lang_i$. The
translation data consists of pairs $(e, f)$ where $e$ is a source
language sentence, and $f$ is a target language sentence. A method
such as GIZA++ is used to derive an alignment between the words in $e$
and $f$, for each sentence pair; the source language parser is used to
parse $e$. Each dependency in $e$ is then potentially transferred
through the alignments to create a dependency in the target sentence
$f$. Once dependencies have been transferred in this way, a dependency
parser can be trained on the dependencies in the target language.

The density-driven approach of \newcite{rasooli-collins:2015:EMNLP} makes use of various
definitions of ``density'' of the projected dependencies. For example,
${\cal P}_{100}$ is the set of projected structures where the
projected dependencies form a full projective parse tree for the
sentence; ${\cal P}_{80}$ is the set of projected structures where at
least 80\% of the words in the projected structure are a modifier in
some dependency. An iterative training process is used, where the
parsing algorithm is first trained on the set ${\cal T}_{100}$ of
complete structures, and where progressively less dense structures are
introduced in learning.

We integrate our approach with the density-driven approach of
\newcite{rasooli-collins:2015:EMNLP} as follows: Consider the treebanks $\tb_1 \ldots \tb_m$
created using the lexicalization method of \S\ref{expansion_section}. We add
all trees in these treebanks to the set ${\cal P}_{100}$ of full trees
used to initialize the method of \newcite{rasooli-collins:2015:EMNLP}. In addition we make
use of the representations $\phi^{(p)}, \phi^{(c)}$ and $\phi^{(l)}$,
throughout the learning
process.

\section{Experiments}
\label{sec_experiments}
This section first describes the experimental settings, then reports results.

\subsection{Data and Tools}
\paragraph{Data}

In a first set of experiments, we consider 7 European languages
studied in several pieces of previous work
\cite{ma-xia:2014:P14-1,yuanregina15,guo2016representation,ammar2016one,lacroix-EtAl:2016:N16-1}.
More specifically, we use the 7 European languages in the Google
universal treebank (v.2; standard data)
\cite{mcdonald-EtAl:2013:Short}. As in previous work, gold
part-of-speech tags are used for evaluation. We use the concatenation
of the treebank training sentences, Wikipedia data and the Bible
monolingual sentences as our monolingual raw text.
Table~\ref{tab_mono_stat} shows statistics for the monolingual data.
We use the Bible data from \newcite{christodouloupoulos2014massively},
which includes data for 100 languages, as the source of
translations. We also conduct experiments with the Europarl data (both
with the original size and a subset of it with the same size as the
Bible text) to study the effects of translation data size and domain
shift. The statistics for translation data is shown in Table
\ref{tab_parallel_stat}.

In a second set of experiments, we ran experiments on 38 datasets (26
languages) in the more recent Universal Dependencies v1.3 corpus
\cite{11234/1-1699}. The full set of languages we use is listed in
Table~\ref{universal_results}.\footnote{We excluded languages that are not completely
  present in the Bible corpus of
  \newcite{christodouloupoulos2014massively} (Ancient Greek, Basque,
  Catalan, Galician, Gothic, Irish, Kazakh, Latvian, Old Church
  Slavonic, and Tamil). We also excluded Arabic, Hebrew, Japanese and
  Chinese, as these languages have tokenization and/or morphological
  complexity that goes beyond the scope of this paper. Future work
  should consider these languages.}  We use the Bible as the
translation data, and Wikipedia as the monolingual text.  The standard
training, development and test set splits are used in all experiments.
The development sets are used for analysis, given  in
\S~\ref{analysis_section} of this paper.

\begin{table}[ht!]
    \centering
    \small
    \setlength{\tabcolsep}{3pt}

    \begin{tabular}{l c c c c c  c c}
    \hline
    Lang. & en & de & es & fr & it & pt & sv \\ \hline
    \#Sen. & 31.8 & 20.0 & 13.6 & 13.6 & 10.1 & 6.1 & 3.9 \\
    \#Token & 750.5 & 408.2 & 402.3 & 372.1 & 311.1 & 169.3 & 60.6 \\ 
    \#Type & 3.8 & 6.1 & 2.7 & 2.4 & 2.1 & 1.6 & 1.3 \\ \hline
    
    \end{tabular}
    \caption{Sizes of the monolingual datasets for each of our languages.
All numbers are in millions.}
    \label{tab_mono_stat}
\end{table}
\begin{table}[t!]
    \centering
    \footnotesize
 \setlength{\tabcolsep}{2pt}
 \begin{tabular}{ l  c  c  c  c  c c  c  c } 
    \hline
        Data & Lang. & en & de & es & fr & it & pt & sv  \\\hline \hline
       \multirow{2}{*}{Bible} & tokens & 1.5M  & 665K & 657K & 732K & 613K & 670K & 696K   \\ \cline{2-9}
        & types  & 16K & 20K & 27K & 22K & 29K & 29K & 23K    \\ \hline \hline
        \multirow{2}{*}{EU-S} & tokens &  718K & 686K & 753K & 799K & 717K & 739K & 645K \\ \cline{2-9}
        & types  &  22K & 41K & 31K & 27K & 30K & 32K &  39K   \\ \hline \hline
          \multirow{2}{*}{Europarl} & tokens & 56M &  50M & 57M & 62M & 55M &  56M &  46M  \\ \cline{2-9}
            & types &  133K & 400K & 195K & 153K &  188K & 200K  & 366K    \\\hline
    \end{tabular}
    \caption{Statistics for the Bible, sampled Europarl (EU-S) and Europarl datasets. Each individual Bible text file from \protect\newcite{christodouloupoulos2014massively}
consists of 24720 sentences, except for English datasets, where two translations into English are available, giving double the
amount of data. Each text file from the sampled Europarl datasets consists of 25K sentences and Europarl has approximately 2 million sentences per language pair.}
    \label{tab_parallel_stat}
\end{table}

\paragraph{Brown Clustering Algorithm}

We use the off-the-shelf Brown clustering
tool\footnote{\url{https://github.com/percyliang/brown-cluster}}
\cite{liang2005semi} to train monolingual Brown clusters with 500
clusters. The monolingual Brown clusters are used as features over
lexicalized values created in $\phi^{(l)}$, and in self-training
experiments. We train our cross-lingual clustering with the
off-the-shelf-tool\footnote{\url{https://github.com/karlstratos/singular}}
from \newcite{stratos-collins-hsu:2015:ACL-IJCNLP}. We set the window
size to 2 with cluster
size of 500.\footnote{Usually the original Brown clusters are better
  features for parsing but their training procedure does not scale well to
  large datasets. Therefore we use the more efficient algorithm from
 \newcite{stratos-collins-hsu:2015:ACL-IJCNLP} on
  the larger cross-lingual datasets to obtain word clusters.}

\paragraph{Parsing Model}

We use the k-beam arc-eager dependency parser of
\newcite{rasooli2015yara}, which is similar to the model of
\newcite{zhang-nivre:2011:ACL-HLT2011}. We modify the parser such that
it can use both monolingual and cross-lingual word cluster
features. The parser is trained using the the maximum violation update
strategy \cite{huang-fayong-guo:2012:NAACL-HLT}. We use three epochs
of training for all experiments. We use the Dependable
tool~\cite{choi2015depends} to calculate significance tests
on several of the comparisons (details are given in the captions
to tables~\ref{tab_results1}, ~\ref{tab_results3}, and ~\ref{universal_results}).

\paragraph{Word alignment}
We use the
intersected alignments from Giza++ \cite{och2000giza} on translation
data. We exclude sentences in translation data with more than $100$ words.

 \subsection{Results on the Google Treebank}
 
Table \ref{tab_results1} shows the dependency parsing accuracy for the
baseline delexicalized approach, and for models which add 1)
cross-lingual clusters (\S\ref{sec:clusters}); 2) lexical features (\S\ref{expansion_section});
3) integration with the density-driven method of \newcite{rasooli-collins:2015:EMNLP}.  It can be
seen that each of these three steps gives significant improvements in
performance. The final LAS/UAS of 73.9/80.3\% is several percentage
points higher than the baseline accuracy of 63.9/72.9\%. 

%MJC: I removed this as the later paragraph talks about table 6
%We also run experiments with a subset of the Europarl data to see the
%effect of domain shift. As shown in Table \ref{tab_results3}, we
%observe that there is a slight drop in average accuracy when we use
%the Bible data instead of Europarl data.

\begin{table}[!t]
\scriptsize
\centering
\setlength{\tabcolsep}{4.5pt}
   \begin{tabular}{c c  c  c  c  c  c c   c  c c}
    \hline
     \multirow{3}{*}{L}   & \multicolumn{2}{c}{\multirow{2}{*}{Baseline}} & ~ & \multicolumn{6}{c}{This paper using the Bible data} \\ \cline{5-10}
       & &   & ~ &  \multicolumn{2}{c}{\S\ref{sec:clusters}}  & \multicolumn{2}{c}{\S\ref{expansion_section}} &   \multicolumn{2}{c}{\S\ref{density}} \\ \cline{2-3}\cline{5-10}
         & LAS & UAS & ~ & LAS & UAS & LAS & UAS & LAS & UAS  \\\cline{1-3}\cline{5-10}
        
        en & 58.2 & 65.5 &~ &   65.0 & 72.3 &  66.3  & 74.0  &  {\bf 70.8}  &  {\bf 76.5}   \\
        de &  49.7 & 59.1 &~ &  51.6	& 59.7  &   54.9  & 62.6  &  {\bf  65.2} & {\bf 72.8 }      \\
        es  &  68.3  & 77.2 &~ &  73.1 & 79.6 & 76.6 & 81.9 &   {\bf 76.7} & { \bf 82.1 }     \\
        fr  & 67.3  & 77.7 &~ &  69.5 & 79.9 & 74.4 &   81.9 &   {\bf 75.8}  &  {\bf 82.2 }   \\
        it  & 69.7  & 79.4 &~ &  71.6 & 80.0 &  74.7 & 82.8  &  { \bf 76.1} & {\bf 83.3}    \\ 
        pt  &  71.5 & 77.5 & ~ &  76.9 & 81.5 & 81.0 & 84.4  &   {\bf 81.3}  & {\bf 84.7}    \\
        sv  & 62.6  & 74.2 &~ &  63.5 & 75.1 &  68.2 & 78.7  &   {\bf 71.2}  & {\bf 80.3} \\ \cline{1-3}\cline{5-10}
        avg &   63.9 & 72.9  &~ &   67.3 &	75.5  &  70.9 & 78.1  &   {\bf 73.9}  &  {\bf 80.3} \\ \hline
    \end{tabular}
    \caption{Performance of different models in this paper; first the
      baseline model, then models trained using the methods described
      in sections \S\ref{sec:clusters}--\ref{density}. %All results
      make use of the Bible as a source of translation data. All
      differences in UAS and LAS are statistically significant with $p
      < 0.001$ using McNemar's test, with the exception of de UAS/LAS
      Baseline vs. 3.1 (i.e., 49.7 vs 51.6 UAS and 59.1 vs 59.7 LAS
      are not significant differences).}
    \label{tab_results1}
\end{table}

\begin{table*}[!th!]
 \scriptsize
 \centering
    \begin{tabular}{ c c c c  c  c  c c c c c c c c c c }
    \hline
         \multirow{3}{*}{Lang.}  &  & \multicolumn{4}{c}{Bible} & &  \multicolumn{4}{c}{Europarl-Sample} & & 
         \multicolumn{4}{c}{Europarl}  \\    \cline{3-6} \cline{8-11} \cline{13-16}
& &  \multicolumn{2}{c}{Density}   &\multicolumn{2}{c}{This Paper }  & &  \multicolumn{2}{c}{Density}   &\multicolumn{2}{c}{This Paper }  & &  \multicolumn{2}{c}{Density}    &\multicolumn{2}{c}{This Paper}  \\  \cline{3-6} \cline{8-11} \cline{13-16}
         &  & LAS & UAS & LAS & UAS & & LAS & UAS & LAS & UAS & & LAS & UAS & LAS & UAS \\   \cline{1-1} \cline{3-6} \cline{8-11} \cline{13-16}
         en &  &  59.1 &  66.4 &  { 70.8} & { 76.5} & &  64.3 & 72.8 & { 70.2} & { 76.2} & &  68.4 & 76.3  &  { 71.1} & { 77.5} \\ 
         de &   &  60.2 & 69.5 & { 65.2} & { 72.8} & &  61.6 & 72.0 &  { 64.9} & { 73.0} & &  73.0 &  79.7 & { 75.6} & { 82.1}  \\ 
         es &  &  70.3 & 76.8 &  { 76.7} &  { 82.1}& &  72.0 & 78.3 & { 76.0} & { 81.5} & & 74.6 & 80.9 &  { 76.6} & { 82.6} \\ 
         fr &  &  69.9 & 76.9 & { 75.8} & { 82.2}& &  71.9 & 79.0 & { 75.7 } & { 82.5 } & &   76.3 & 82.7 & { 77.4} &  { 83.9} \\ 
         it &   & 71.1 & 78.5 & { 76.1} & { 83.3}& &  73.2 & 80.4 & { 76.2} & { 82.9}  & & 77.0 &  83.7 & { 77.4} & { 84.4} \\ 
         pt &   & 72.1 & 76.4 &  { 81.3} & { 84.7}& &  75.3 & 79.7 & { 81.61} & { 84.8}  & & 77.3 & 82.1 & { 82.1} & { 85.6} \\  
         sv &   &  66.5 & 76.3 & { 71.2} & { 80.3}& &  71.9 & 80.6 & { 73.5} & { 81.6}  & &  75.6 &  84.1 & { 76.9} & { 84.5} \\   \cline{1-1} \cline{3-6} \cline{8-11} \cline{13-16}
         avg &   & 67.0 & 75.7 & { 73.9} & { 80.3}& &  70.0 & 77.6 & { 74.0} & { 80.4} & &  74.6 & 81.3  & { 76.7} & { 82.9} \\  \hline
    \end{tabular}
    \caption{\footnotesize Results for our method using different
      sources of translation data.  ``Density'' refers to the method
      of \protect\newcite{rasooli-collins:2015:EMNLP}; ``This paper''
      gives results using the methods described in sections
      \ref{sec:clusters}--\ref{density} of this paper.  The ``Bible''
      experiments use the Bible data of
      \protect\newcite{christodouloupoulos2014massively}.  The
      ``Europarl'' experiments use the Europarl data of
      \protect\newcite{koehn2005europarl}.  The ``Europarl-Sample''
      experiments use 25K randomly chosen sentences from Europarl;
      this gives a similar number of sentences to the Bible data. All
      differences in LAS and UAS in this table between the density and
      ``this paper'' settings (i.e., for the Bible, Europarl-Sample
      and Europarl settings) are found to be statistically
      significant according to McNemar's sign test. }
    \label{tab_results3}
\end{table*}

\begin{table*}[ht!]
    \centering
    \scriptsize
    \begin{tabular}{ c  c  c c c  c c c   c  c  c  c c  c  c c c}
    \hline
          \multirow{3}{*}{Lang.} & \multicolumn{1}{c}{ \multirow{2}{*}{MX14}}  & \multicolumn{1}{c}{ \multirow{2}{*}{LA16}} & \multicolumn{2}{c}{ \multirow{2}{*}{ZB15}}  & \multicolumn{2}{c}{\multirow{2}{*}{GCY16}}  & \multicolumn{2}{c}{\multirow{2}{*}{AMB16}} & \multicolumn{2}{c}{\multirow{2}{*}{RC15}} & \multicolumn{4}{c}{This paper} & \multicolumn{2}{c}{\multirow{2}{*}{Supervised}} \\\cline{12-15}
          & & &  & & &  & & &  & & \multicolumn{2}{c}{Bible} &  \multicolumn{2}{c}{Europarl} & & \\ \cline{2-17}
        & UAS & UAS & LAS & UAS& LAS & UAS & LAS & UAS & LAS & UAS &  LAS & UAS& LAS & UAS\\ \hline
         en &   -- & -- & 59.8 &	70.5  & -- & -- & -- & -- & 68.4 & 76.3 &  70.8 &  76.5 & {\bf 71.1} & {\bf 77.5} & 92.0 & 93.8 \\ 
         de & 74.3 & 76.0 & 54.1 & 62.5 & 55.9 & 65.0 & 57.1 & 65.2 & 73.0 & 79.7 &  65.2&  72.8 & {\bf 75.6}	& {\bf 82.1}  & 79.4 & 85.3 \\ 
         es & 75.5 & 78.9 & 68.3 & 78.0 & 73.0 & 79.0 & 74.6 & 80.2 & 74.6 & 80.9 &  {\bf 76.7} &   82.1 & 76.0 & {\bf 82.6} &  82.3 & 86.7 \\ 
         fr & 76.5 & 80.8 & 68.8 & 78.9 & 71.0 & 77.6 & 73.9 & 80.6 & 76.3 & 82.7 &  75.8 &   82.2 & {\bf 77.4} & {\bf 83.9}  & 81.7 & 86.3 \\ 
         it & 77.7 & 79.4 & 69.4 & 79.3 & 71.2 & 78.4 & 72.5 & 80.7 & 77.0 & 83.7 &  76.1 &  83.3 & {\bf 77.4}	& {\bf 84.4}   & 86.1 & 88.8 \\ 
         pt & 76.6 & -- & 72.5 & 78.6 & 78.6 & 81.8 & 77.0 & 81.2 & 77.3  & 82.1 &   81.3  &  84.7 & {\bf 82.1} & {\bf 85.6} & 87.6 &  89.4 \\ 
         sv & 79.3 & 83.0 & 62.5 & 75.0 & 69.5 & 78.2 & 68.1 & 79.0 & 75.6 & 84.1 &  71.2  &   80.3  & {\bf 76.9} & {\bf 84.5}  & 84.1 & 88.1  \\ 
       \hline 
       avg$_{ \text{\textbackslash en}}$ & 76.7 & -- & 65.9 & 75.4 & 69.3 & 76.3 & 70.5 & 77.8 & 75.6 & 82.2 &   74.4&   80.9 & {\bf 77.7} & {\bf 83.9}  & 83.5 &  87.4 \\ \hline 
    \end{tabular}
    \caption{\footnotesize Comparison of our work using Bible and Europarl data,  with previous work: MX14 \protect\cite{ma-xia:2014:P14-1}, LA16 \protect\cite{lacroix-EtAl:2016:N16-1}, ZB15 
\protect\cite{yuanregina15},  
GCY16 
\protect\cite{guo2016representation},  AMB 16
\protect\cite{ammar2016one}, and RC15 \protect\cite{rasooli-collins:2015:EMNLP}. 
``Supervised'' refers to the performance of the parser trained on fully gold standard data in a supervised fashion (i.e. the practical upper-bound of our model). ``avg$_{ \text{\textbackslash en}}$'' refers to the average accuracy for all datasets except English.}
    \label{tab_results2}
\end{table*}

\paragraph{Comparison to the Density-Driven Approach using Europarl Data}
Table \ref{tab_results3} shows accuracies for the density-driven
approach of \newcite{rasooli-collins:2015:EMNLP}, first using Europarl
data\footnote{\newcite{rasooli-collins:2015:EMNLP} do not report
  results on English. We use the same setting as in their paper to
  obtain the English results.} and second using the Bible data alone
(with no cross-lingual clusters or lexicalization). The Bible data is
considerably smaller than Europarl (around 100 times smaller), and it
can be seen that results using the Bible are several percentage points
lower than the results for Europarl (75.7\% UAS vs. 81.3\%
UAS). Integrating cluster and lexicalized features described in the
current paper with the density-driven approach closes much of this gap
in performance (80.3\% UAS). Thus we have demonstrated that we can
get close to the performance of the Europarl-based models using only
the Bible as a source of translation data. Using our approach on
the full Europarl data gives an average UAS of 82.9\%, an improvement
from the 81.3\% UAS of \newcite{rasooli-collins:2015:EMNLP}. 

%Our method improves more over the density-driven method when the
%translation data is smaller. This is mostly due to the fact that when
%translation data is large enough, we can extract more accurate
%projected trees leading to better parsing performance.

Table~\ref{tab_results3} also shows results when we use a random subset of the Europarl data, in which the number of sentences (25,000) is chosen to
give a very similar size to the Bible dataset. It can be seen that
accuracies using the Bible data vs. Europarl-Sample are very similar
(80.3\% vs. 80.4\% UAS), suggesting that the size of the translation
corpus is much more important than the genre.

\paragraph{Comparison to Other Previous Work}
Table \ref{tab_results2} compares the accuracy of our method to the
following related work: 1) \newcite{ma-xia:2014:P14-1}, who describe an annotation projection method based on entropy regularization; 2) \newcite{lacroix-EtAl:2016:N16-1}, who describe an annotation projection method based on training on partial trees with dynamic oracles; 3) \newcite{yuanregina15}, who describe a method that learns
cross-lingual embeddings and bilingual dictionaries from Europarl
data, and uses these features in a discriminative parsing model; 4)
\newcite{guo2016representation}, who describe a method
that learns cross-lingual embeddings from Europarl data and
uses a shift-reduce neural parser with these representations; 5)
\newcite{ammar2016one}, who use the same embeddings as 
\newcite{guo2016representation}, within an LSTM-based
parser; and 6) \newcite{rasooli-collins:2015:EMNLP} who use the density-driven approach on the Europarl data. Our method gives significant improvements over the first three models, in spite of using the Bible translation data rather than Europarl. When using the Europarl data, our method improves the state-of-the-art model of \newcite{rasooli-collins:2015:EMNLP}. %It is worth noting that our model significantly outperform the UAS reported in other previous work \cite{mcdonald-petrov-hall:2011:EMNLP,ma-xia:2014:P14-1,lacroix-EtAl:2016:N16-1} but we do not put theirs because of space restrictions.

\begin{table}[t!]
  \scriptsize
\centering
    \begin{tabular}{ c  c c    c c   c c }
    \hline
         \multirow{3}{*}{Lang.}  &  \multicolumn{2}{c}{\multirow{2}{*}{RC15}}   &\multicolumn{4}{c}{This Paper (\S\ref{density})} \\ \cline{4-7}
         & &  &\multicolumn{2}{c}{Bible}  & \multicolumn{2}{c}{Europarl}  \\  
         \cline{2-7}
         & LAS & UAS & LAS & UAS & LAS & UAS  \\ \hline
en & 66.2 & 74.4 & 67.8 & 74.4 & {\bf 68.0} & {\bf 75.1}  \\ 
 de & 71.6 & 78.8 & 61.9 & 70.3 & {\bf 73.6} & {\bf 80.8}  \\ 
 es & 72.3 & 79.2 & 73.8 & 79.9 & {\bf 74.2} & {\bf 80.7}  \\ 
 fr & 73.5 & 80.8 & 72.6 & 79.9 & {\bf 75.0} & {\bf 82.3}  \\ 
 it & 74.9 & 82.0 & 74.0 & 81.7 & {\bf 75.3} & {\bf 82.6}  \\ 
 pt & 75.4 & 80.7 & 79.2 & 83.3& {\bf 80.4} & {\bf 84.4}  \\ 
 sv & 73.4 & 82.0 & 67.3 & 77.2& {\bf 73.7} & {\bf 82.2}  \\ \hline
 avg & 72.5 & 79.7 & 70.9 & 78.1 & {\bf 74.3} & {\bf 81.2}  \\ \hline
    \end{tabular}
    \caption{\footnotesize  The final results based on automatic part of speech tags. RC15 refers to the best performing model of \protect\newcite{rasooli-collins:2015:EMNLP}.}
    \label{tab_auto_results}
    
\end{table}

\begin{table}[!th]
    \centering
    \scriptsize
    \setlength{\tabcolsep}{4.5pt}
    \begin{tabular}{ l | c c | c c | c c }
    \hline \hline  
      \multirow{2}{*}{Dataset}   & \multicolumn{2}{c}{Density}  &  \multicolumn{2}{c}{This paper}  &  \multicolumn{2}{c}{Supervised} \\  \cline{2-7} 
    &  LAS  &  UAS  &  LAS  &  UAS  &  LAS &  UAS  \\ \hline 
it & 74.3 & 81.3 & 79.8 & 86.1 & 88.4 & 90.7 \\ 
sl & 68.2 & 75.9 & 78.6 & 84.1 & 86.3 & 89.1 \\ 
es & 69.1 & 77.5 & 76.3 & 84.1 & 83.5 & 86.9 \\ 
bg & 66.2 & 79.5 & 72.0 & 83.6 & 85.5 & 90.5 \\ 
pt & 66.7 & 75.8 & 74.8 & 83.4 & 83.0 & 86.7 \\ 
es-ancora & 68.9 & 77.5 & 74.6 & 83.1 & 86.5 & 89.4 \\ 
fr & 72.0 & 77.9 & 76.6 & 82.6 & 84.5 & 87.1 \\ 
sv-lines & 67.5 & 76.7 & 73.3 & 82.4 & 81.0 & 85.4 \\ 
pt-br & 68.3 & 75.2 & 76.2 & 82.0 & 87.8 & 89.7 \\ 
sv & 65.9 & 75.7 & 71.7 & 81.3 & 83.6 & 87.7 \\ 
no & 71.7 & 78.8 & 74.3 & 81.2 & 88.0 & 90.5 \\ 
pl & 65.4 & 77.6 & 70.1 & 81.0 & 85.1 & 90.3 \\ 
hr & 55.8 & 70.2 & 65.9 & 80.9 & 76.2 & 85.1 \\ 
cs-cac & 61.1 & 70.3 & 69.0 & 78.5 & 82.4 & 87.6 \\ 
da & 63.1 & 72.8 & 68.3 & 77.8 & 80.8 & 84.3 \\ 
en-lines & 67.0 & 75.9 & 68.6 & 77.3 & 80.7 & 84.6 \\ 
cs & 59.0 & 68.1 & 67.2 & 76.4 & 84.5 & 88.7 \\ 
id & 38.0 & 55.7 & 57.8 & 76.0 & 79.8 & 85.1 \\ 
de & 61.3 & 72.8 & 64.9 & 75.7 & 80.2 & 85.8 \\ 
ru-syntagrus & 56.0 & 70.7 & 61.6 & 75.3 & 82.0 & 87.8 \\ 
ru & 56.7 & 64.8 & 65.4 & 74.8 & 71.9 & 77.7 \\ 
cs-cltt & 57.5 & 65.4 & 65.6 & 74.7 & 77.1 & 81.4 \\ 
ro & 54.6 & 67.4 & 60.7 & 74.6 & 78.2 & 85.3 \\ 
la & 54.5 & 71.6 & 55.7 & 72.8 & 43.1 & 52.5 \\ 
nl-lassysmall & 51.5 & 62.6 & 61.9 & 71.7 & 76.5 & 80.6 \\ 
el & 53.7 & 66.7 & 59.6 & 71.0 & 79.1 & 83.1 \\ 
et & 48.9 & 65.6 & 56.9 & 70.9 & 75.9 & 82.9 \\ 
hi & 34.4 & 50.6 & 49.9 & 69.9 & 89.4 & 92.9 \\ 
hu & 26.1 & 48.9 & 55.0 & 69.9 & 69.5 & 79.4 \\ 
en & 59.7 & 68.1 & 61.8 & 69.0 & 85.3 & 88.1 \\ 
fi-ftb & 50.3 & 63.2 & 56.5 & 67.5 & 73.3 & 79.7 \\ 
fi & 49.8 & 60.8 & 57.3 & 66.4 & 73.4 & 78.2 \\ 
la-ittb & 44.1 & 55.4 & 51.8 & 62.8 & 76.2 & 80.9 \\ 
nl & 40.6 & 49.4 & 50.1 & 62.0 & 70.1 & 75.0 \\ 
la-proiel & 43.6 & 60.3 & 45.0 & 61.3 & 64.9 & 72.9 \\ 
sl-sst & 42.4 & 59.2 & 47.6 & 60.6 & 63.4 & 70.4 \\ 
fa & 44.4 & 53.2 & 46.5 & 56.0 & 84.1 & 87.5 \\ 
tr & 05.3 & 18.5 & 32.7 & 51.9 & 65.6 & 78.8 \\  \hline
Average & 56.7 & 68.1 & 64.0 & 74.8 & 78.9 & 83.8 \\ \hline  \hline
    \end{tabular}
    \caption{\footnotesize Results for the density driven method \protect\cite{rasooli-collins:2015:EMNLP} and ours using the Bible data on the universal dependencies v1.3 \protect\cite{11234/1-1699}. The table is sorted by the performance of our method. The last major columns shows the performance of the supervised parser. The abbreviations are as follows: bg (Bulgarian), cs (Czech), da (Danish), de (German), el (Greek), en (English), es (Spanish), et (Estonian), fa (Persian (Farsi)), fi (Finnish), fr (French), hi (Hindi), hr (Croatian), hu (Hungarian), id (Indonesian), it (Italian), la (Latin), nl (Dutch), no (Norwegian), pl (Polish), pt (Portuguese), ro (Romanian), ru (Russian), sl (Slovenian), sv (Swedish), and tr (Turkish). All differences in LAS and UAS in this table were found to be statistically significant according to McNemar's sign test with $p < 0.001$. }
    \label{universal_results}
\end{table}

\paragraph{Performance with Automatic POS Tags} 

For completeness, Table~\ref{tab_auto_results} gives results for our
method with automatic part-of-speech tags. The tags are obtained using
the model of \newcite{collins:2002:EMNLP02}\footnote{\url{https://github.com/rasoolims/SemiSupervisedPosTagger}} trained
on the training part of the treebank dataset.
Future work should study approaches that transfer POS tags in
addition to dependencies.

%Although there is a drop in accuracy when using automatic
%POS tags, our model has still a high accuracy even higher than those
%from the density driven approach of
%\newcite{rasooli-collins:2015:EMNLP}.

%\footnote{Future work
%  should study the use of transferred tags instead of supervised
%  tags.} 

%\input{good_vs_1}

\subsection{Results on the Universal Dependencies v1.3}
\label{sec_univ_res}
Table~\ref{universal_results} gives results on 38 datasets (26 languages) from the
newly released universal dependencies corpus
\cite{11234/1-1699}. Given the number of treebanks and to speed up
training, we pick source languages that have at least 5 out of 6
common WALS properties with each target language. Our experiments are
carried out using the Bible as our translation data.  As shown in
Table \ref{universal_results}, our method consistently outperforms the
density-driven method of \newcite{rasooli-collins:2015:EMNLP} and for
many languages the accuracy of our method gets close to the accuracy
of the supervised parser. In all the languages, our method is significantly better than the density-driven method using the McNemar's test with $p<0.001$.

Accuracy on some languages (e.g., Persian
(fa) and Turkish (tr)) is low, suggesting that future work should
consider more powerful techniques for these languages. There are two important facts to note. First, the number of fully projected trees in some languages is so low such that the density-driven approach cannot start with a good initialization to fill in partial dependencies. For example Turkish has only one full tree with only six words, Persian with 25 trees, and Dutch with 28 trees. Second, we observe very low accuracies in supervised parsing for some languages in which the number of training sentences is very low (for example, Latin has only 1326 projective trees in the training data).

\begin{table}[!t]
    \centering
    \begin{scriptsize}
    \setlength{\tabcolsep}{1pt}
 
    \begin{tabular}{l|c | c c | c  c | c c}
    \hline \hline
     \multirow{2}{*}{dependency} & \multirow{2}{*}{freq}   & \multicolumn{2}{c|}{Delexicalized}  &  \multicolumn{2}{c|}{Bible} & \multicolumn{2}{c}{Europarl} \\ 
          &  & prec./rec. & f1 & prec./rec. & f1 & prec./rec. & f1   \\ \hline
adpmod & 10.6 & 57.2/62.7 & 59.8 & 67.1/71.8 & 69.4 & 70.3/73.8 & 72.0 \\ 
adpobj & 10.6 & 65.5/69.1 & 67.2 & 75.3/77.4 & 76.3 & 75.9/79.2 & 77.6 \\ 
det & 9.5 & 72.5/75.6 &74.0 & 84.3/86.3 & 85.3 & 86.6/89.8 & 88.2 \\ 
compmod & 9.1 & 83.7/ 59.9 &69.8 & 87.3/70.2 & 77.8 & 89.0/73.0 & 80.2 \\ 
nsubj & 8.0 & 69.7/60.0 & 64.5 & 82.1/77.5 & 79.7 & 83.0/78.1 & 80.5 \\ 
amod & 7.0 & 76.9/72.3 & 74.5 & 83.0/78.7 & 80.8 & 80.9/77.9 & 79.4 \\ 
ROOT & 4.8 & 69.3/70.4 & 69.8 & 85.0/85.1 & 85.0 & 83.8/85.8 & 84.8 \\ 
num & 4.6 & 67.8/55.3 & 60.9 & 70.7/55.2 & 62.0 & 75.0/63.0 & 68.5 \\ 
dobj & 4.5 & 60.8/80.3 & 69.2 & 64.0/84.9 & 73.0 & 68.4/86.6 & 76.5 \\ 
advmod & 4.1 & 65.9/61.9 & 63.8 & 72.7/68.1 & 70.3 & 69.6/68.8 & 69.2 \\ 
aux & 3.5 & 76.6/93.9 & 84.4 & 90.2/95.9 & 93.0 & 89.6/96.4 & 92.9 \\ 
cc & 2.9 & 67.6/61.7 & 64.5 & 73.1/73.1 & 73.1 & 73.1/73.3 & 73.2 \\ 
conj & 2.8 & 46.3/56.1 & 50.7 & 45.6/62.9 & 52.9 & 48.1/62.8 & 54.5 \\ 
dep & 2.0 & 90.5/25.8 & 40.1 & 99.2/33.8 & 50.4 & 92.0/34.4 & 50.1 \\ 
poss & 2.0 & 72.1/30.6 & 43.0 & 77.9/45.8 & 57.7 & 78.2/42.1 & 54.7 \\ 
ccomp & 1.6 & 76.2/28.4 & 41.3 & 88.0/61.3 & 72.3 & 82.3/69.1 & 75.1 \\ 
adp & 1.2 & 20.0/0.5 & 0.9 & 92.7/42.1 & 57.9 & 91.7/23.3 & 37.1 \\ 
nmod & 1.2 & 60.7/48.1 & 53.7 & 56.3/47.1 & 51.3 & 52.6/46.2 & 49.2 \\ 
xcomp & 1.2 & 66.6/48.6 & 56.2 & 85.1/65.3 & 73.9 & 78.3/71.0 & 74.5 \\ 
mark & 1.1 & 37.8/24.6 & 29.8 & 73.8/50.3 & 59.8 & 62.8/53.8 & 57.9 \\ 
advcl & 0.8 & 23.6/22.3 & 22.9 & 38.7/38.8 & 38.8 & 38.0/42.9 & 40.3 \\ 
appos & 0.8 & 8.5/43.0 & 14.3 & 20.4/61.0 & 30.6 & 26.4/61.7 & 37.0 \\ 
auxpass & 0.8 & 88.9/91.4 & 90.1 & 96.8/97.1 & 97.0 & 98.6/98.6 & 98.6 \\ 
rcmod & 0.8 & 38.2/33.3 & 35.6 & 46.8/54.6 & 50.4 & 52.7/55.0 & 53.8 \\ 
nsubjpass & 0.7 & 73.2/64.9 & 68.8 & 87.6/77.0 & 82.0 & 85.5/75.8 & 80.3 \\ 
acomp & 0.6 & 86.8/92.5 & 89.6 & 83.3/93.5 & 88.1 & 91.0/93.9 & 92.4 \\ 
adpcomp & 0.6 & 42.0/70.2 & 52.5 & 47.9/61.5 & 53.9 & 55.4/47.1 & 50.9 \\ 
partmod & 0.6 & 20.2/36.0 & 25.8 & 36.7/49.1 & 42.0 & 31.0/40.7 & 35.2 \\ 
attr & 0.5 & 67.7/86.4 & 75.9 & 76.5/92.1 & 83.6 & 72.6/92.7 & 81.4 \\ 
neg & 0.5 & 74.7/85.0 & 79.6 & 93.3/91.0 & 92.1 & 92.6/89.8 & 91.2 \\ 
prt & 0.3 & 27.4/92.2 & 42.2 & 32.4/96.6 & 48.5 & 31.9/97.4 & 48.1 \\ 
infmod & 0.2 & 30.7/72.4 & 43.2 & 38.4/64.4 & 48.1 & 42.6/63.2 & 50.9 \\ 
expl & 0.1 & 84.8/87.5 & 86.2 & 93.8/93.8 & 93.8 & 91.2/96.9 & 93.9 \\ 
iobj & 0.1 & 51.7/78.9 & 62.5 & 88.9/84.2 & 86.5 & 36.4/84.2 & 50.8 \\ 
mwe & 0.1 & 0.0/0.0 & 0.0 & 5.3/2.1 & 3.0 & 11.1/10.4 & 10.8 \\ 
parataxis & 0.1 & 5.6/19.6 & 8.7 & 17.3/47.1 & 25.3 & 14.6/45.1 & 22.0 \\ 
cop & 0.0 & 0.0/0.0 & 0.0 & 0.0/0.0 & 0.0 & 0.0/0.0 & 0.0 \\ 
csubj & 0.0 & 12.8/33.3 & 18.5 & 22.2/26.7 & 24.2 & 25.0/46.7 & 32.6 \\ 
csubjpass & 0.0 & 100.0/100.0 & 100.0 & 100.0/100.0 & 100.0 & 50.0/100.0 & 66.7 \\ 
rel & 0.0 & 100.0/6.3 & 11.8 & 90.9/62.5 & 74.1 & 66.7/37.5 & 48.0\\ \hline \hline
    \end{tabular}
        \caption{\footnotesize Precision, recall and f-score of different dependency relations on the English development data of the Google universal treebank. The major columns show the dependency labels (``dep.''), frequency (``freq.''), the baseline delexicalized model (``delex''), and our method using the Bible and Europarl (``EU'') as translation data. The rows are sorted by frequency.}
    \label{tab_en_dep_las}
    \end{scriptsize}
\end{table}

\begin{table}[!ht]
    \centering
    \begin{scriptsize}

    \begin{tabular}{l|c c  |c c | c c}
    \hline \hline
        \multirow{2}{*}{POS} & \multicolumn{2}{c|}{{\cal G}1 }  &  \multicolumn{2}{c|}{{\cal G}2 }  & \multicolumn{2}{c}{{\cal G}3 } \\
         & freq\% & acc. & freq\% & acc. & freq\% & acc.  \\ \hline
         NOUN & 22.0 & 77.6 & 30.0 & 71.2 & 25.3 & 58.0 \\ 
ADP & 16.9 & 92.3 & 10.9 & 92.3 & 11.2 & 90.6 \\ 
DET & 11.9 & 96.4 & 3.0 & 92.4 & 3.6 & 86.6 \\ 
VERB & 11.7 & 74.5 & 13.5 & 66.1 & 17.1 & 52.2 \\ 
PROPN & 8.1 & 79.0 & 4.7 & 65.2 & 6.8 & 49.5 \\ 
ADJ & 8.0 & 88.5 & 12.7 & 86.9 & 8.4 & 73.6 \\ 
PRON & 5.4 & 87.7 & 5.9 & 82.2 & 7.6 & 71.1 \\ 
ADV & 4.3 & 76.0 & 6.6 & 70.9 & 5.6 & 61.9 \\ 
CONJ & 3.6 & 71.8 & 4.7 & 63.0 & 4.2 & 60.4 \\ 
AUX & 2.7 & 91.5 & 1.7 & 88.9 & 3.0 & 70.6 \\ 
NUM & 2.2 & 79.5 & 2.3 & 68.4 & 2.0 & 75.7 \\ 
SCONJ & 1.8 & 80.5 & 1.9 & 77.2 & 2.6 & 65.0 \\ 
PART & 0.9 & 80.2 & 1.8 & 64.3 & 1.9 & 45.0 \\ 
X & 0.2 & 52.3 & 0.1 & 40.5 & 0.6 & 36.9 \\ 
SYM & 0.1 & 64.3 & 0.1 & 40.9 & 0.1 & 45.5 \\ 
INTJ & 0.1 & 78.5 & 0.0 & 51.7 & 0.3 & 60.2 \\ 
      \hline \hline
    \end{tabular}
    \caption{\footnotesize Accuracy of unlabeled dependencies by POS of the modifier word, for three groups of languages for the universal dependencies experiments in Table~\protect\ref{universal_results}: G1 (languages with $\text{UAS}\geq 80$), G2 (languages with $70\leq \text{UAS} <80$), G3 (languages with $\text{UAS}<70$). The rows are sorted by frequency in the G1 languages.}
    \label{tab:pos_dep_accuracy}
    \end{scriptsize}
\end{table}

\begin{table*}[!ht]
\begin{scriptsize}
\centering
    \begin{tabular}{l | c c c c | c c c c | c c c c}
       \hline \hline
    \multirow{2}{*}{POS} & \multicolumn{4}{c|}{{\cal G}1 }  &  \multicolumn{4}{c|}{{\cal G}2 }  & \multicolumn{4}{c}{{\cal G}3} \\
      & freq\% & prec. & rec. & f1 & freq\% &  prec. & rec. & f1  & freq\% &  prec. & rec. & f1   \\ \hline
NOUN & 43.9 & 85.4 & 88.6 & 87.0 & 43.5 & 77.3 & 81.2 & 79.2 & 34.5 & 67.1 & 71.0 & 69.0 \\ 
VERB & 32.0 & 83.5 & 83.6 & 83.6 & 35.4 & 74.9 & 77.9 & 76.4 & 41.3 & 63.8 & 66.5 & 65.1 \\ 
PROPN & 9.1 & 84.0 & 84.0 & 84.0 & 4.1 & 67.6 & 63.2 & 65.3 & 6.4 & 57.2 & 54.8 & 56.0 \\ 
ADJ & 4.5 & 76.2 & 72.4 & 74.3 & 5.7 & 75.7 & 56.0 & 64.4 & 5.8 & 64.9 & 49.1 & 55.9 \\ 
PRON & 1.4 & 79.3 & 68.3 & 73.4 & 1.4 & 81.5 & 61.4 & 70.0 & 2.2 & 65.2 & 49.1 & 56.0 \\ 
NUM & 1.2 & 77.2 & 72.4 & 74.7 & 1.0 & 52.0 & 41.8 & 46.3 & 0.7 & 62.5 & 54.7 & 58.3 \\ 
ADV & 1.0 & 54.0 & 39.0 & 45.3 & 1.5 & 56.5 & 27.2 & 36.7 & 1.2 & 44.1 & 25.8 & 32.6 \\ 
ADP & 0.6 & 39.8 & 6.5 & 11.2 & 0.3 & 25.0 & 0.9 & 1.7 & 0.3 & 40.5 & 8.3 & 13.8 \\ 
SYM & 0.3 & 79.0 & 81.1 & 80.1 & 0.1 & 41.5 & 66.3 & 51.0 & 0.1 & 55.3 & 52.2 & 53.7 \\ 
DET & 0.3 & 36.3 & 22.6 & 27.8 & 0.1 & 60.6 & 30.6 & 40.7 & 0.1 & 67.6 & 25.3 & 36.8 \\ 
AUX & 0.2 & 35.7 & 3.7 & 6.6 & 0.0 & 17.2 & 6.7 & 9.6 & 0.8 & 33.3 & 2.2 & 4.2 \\ 
X & 0.1 & 52.4 & 52.2 & 52.3 & 0.1 & 42.5 & 41.6 & 42.1 & 0.4 & 39.7 & 42.7 & 41.1 \\ 
SCONJ & 0.1 & 36.8 & 10.0 & 15.7 & 0.1 & 45.7 & 5.8 & 10.3 & 0.1 & 30.0 & 13.5 & 18.7 \\ 
PART & 0.1 & 26.7 & 3.0 & 5.4 & 0.1 & 15.9 & 4.3 & 6.8 & 0.1 & 26.7 & 36.8 & 30.9 \\ 
CONJ & 0.1 & 47.8 & 6.5 & 11.4 & 0.1 & 3.3 & 0.9 & 1.4 & 0.1 & 51.7 & 10.2 & 17.0 \\ 
INTJ & 0.0 & 52.4 & 47.8 & 50.0 & 0.0 & 20.0 & 7.1 & 10.5 & 0.1 & 44.2 & 43.0 & 43.6 \\\hline \hline 
    \end{tabular}
     \caption{\footnotesize Precision, recall and f-score of unlabeled dependency attachment for different POS tags \emph{as head} for three groups of languages for the universal dependencies experiments in Table~\protect\ref{universal_results}: G1 (languages with $\text{UAS}\geq 80$), G2 (languages with $70\leq \text{UAS} <80$), G3 (languages with $\text{UAS}<70$). The rows are sorted by frequency in the G1 languages.}
    \label{tab:pos_head_accuracy}
    \end{scriptsize}
\end{table*}

\begin{table*}[!ht]
\begin{scriptsize}
\centering
 \begin{tabular}{l | c c c c | c c c c | c c c c}
       \hline \hline
    \multirow{2}{*}{Dep.} & \multicolumn{4}{c|}{{\cal G}1 }  &  \multicolumn{4}{c|}{{\cal G}2 }  & \multicolumn{4}{c}{{\cal G}3} \\
      & freq\% & prec. & rec. & f1 & freq\% &  prec. & rec. & f1  & freq\% &  prec. & rec. & f1   \\ \hline    
   nmod & 15.8 & 74.0 & 76.3 & 75.2 & 16.4 & 67.3 & 72.2 & 69.7 & 17.3 & 56.9 & 57.6 & 57.3 \\ 
case & 15.3 & 92.6 & 94.7 & 93.7 & 10.7 & 92.4 & 93.5 & 93.0 & 10.7 & 90.2 & 90.2 & 90.2 \\ 
det & 11.8 & 96.5 & 96.4 & 96.4 & 3.5 & 91.8 & 91.9 & 91.9 & 3.8 & 79.1 & 86.4 & 82.6 \\ 
nsubj & 6.5 & 85.3 & 86.8 & 86.0 & 7.5 & 75.5 & 73.5 & 74.5 & 7.8 & 61.0 & 63.2 & 62.1 \\ 
amod & 6.4 & 92.9 & 94.0 & 93.5 & 10.8 & 90.1 & 90.9 & 90.5 & 5.3 & 75.7 & 82.9 & 79.1 \\ 
dobj & 5.3 & 93.0 & 90.8 & 91.9 & 7.1 & 84.3 & 81.8 & 83.0 & 5.7 & 71.9 & 72.6 & 72.3 \\ 
root & 5.3 & 84.8 & 85.2 & 85.0 & 6.8 & 77.5 & 77.9 & 77.7 & 7.9 & 64.9 & 65.7 & 65.3 \\ 
advmod & 4.1 & 73.4 & 72.2 & 72.8 & 7.1 & 68.1 & 69.3 & 68.7 & 5.3 & 54.8 & 58.7 & 56.7 \\ 
conj & 4.0 & 60.4 & 68.1 & 64.0 & 5.8 & 50.2 & 56.6 & 53.2 & 4.2 & 41.3 & 48.1 & 44.5 \\ 
cc & 3.4 & 71.2 & 71.2 & 71.2 & 4.5 & 63.5 & 63.3 & 63.4 & 3.9 & 60.6 & 61.6 & 61.1 \\ 
mark & 3.3 & 85.1 & 87.0 & 86.0 & 2.2 & 76.2 & 79.6 & 77.9 & 3.4 & 70.9 & 71 & 71 \\ 
acl & 2.4 & 65.9 & 61.6 & 63.7 & 1.7 & 49.7 & 51.3 & 50.5 & 2.0 & 32.6 & 28.7 & 30.5 \\ 
aux & 2.2 & 91.5 & 93.6 & 92.5 & 1.2 & 86.8 & 91.1 & 88.9 & 2.2 & 66.4 & 78.2 & 71.8 \\ 
name & 1.9 & 86.5 & 86.2 & 86.4 & 1.3 & 75.3 & 72.1 & 73.6 & 0.8 & 27.8 & 45.1 & 34.4 \\ 
cop & 1.6 & 73.1 & 74.5 & 73.8 & 1.3 & 67.7 & 52.5 & 59.1 & 2.1 & 50.8 & 51.2 & 51 \\ 
nummod & 1.4 & 83.8 & 86.0 & 84.9 & 1.6 & 73.9 & 77.6 & 75.7 & 1.4 & 79.2 & 81.7 & 80.5 \\ 
advcl & 1.3 & 60.1 & 59.8 & 60.0 & 1.3 & 57.4 & 48.8 & 52.7 & 2.0 & 42.6 & 38.1 & 40.2 \\ 
appos & 1.3 & 73.9 & 64.9 & 69.1 & 0.8 & 51.2 & 48.9 & 50.0 & 0.5 & 31.3 & 32.1 & 31.7 \\ 
mwe & 0.9 & 57.7 & 15.6 & 24.6 & 0.5 & 66.2 & 15.1 & 24.6 & 0.3 & 31.9 & 15.6 & 20.9 \\ 
xcomp & 0.8 & 82.9 & 74.6 & 78.6 & 1.2 & 76.2 & 73.4 & 74.8 & 1.0 & 40.7 & 62.9 & 49.5 \\ 
ccomp & 0.8 & 72.8 & 70.8 & 71.8 & 0.6 & 63.1 & 64.1 & 63.6 & 1.2 & 42.8 & 40.3 & 41.5 \\ 
neg & 0.7 & 89.5 & 88.1 & 88.8 & 0.7 & 81.2 & 82.1 & 81.6 & 1.1 & 73.6 & 72 & 72.8 \\ 
iobj & 0.7 & 98.7 & 91.1 & 94.7 & 0.5 & 96.3 & 71.0 & 81.7 & 1.1 & 97.1 & 67.1 & 79.3 \\ 
expl & 0.6 & 90.9 & 84.7 & 87.7 & 0.7 & 87.3 & 86.8 & 87.1 & 0.1 & 62.5 & 45 & 52.3 \\ 
auxpass & 0.5 & 95.7 & 96.5 & 96.1 & 0.7 & 98.3 & 93.5 & 95.8 & 1.2 & 92.3 & 49.8 & 64.7 \\ 
nsubjpass & 0.5 & 94.6 & 89.9 & 92.2 & 0.7 & 96.1 & 85.0 & 90.2 & 0.6 & 94.4 & 67.2 & 78.5 \\ 
parataxis & 0.4 & 56.0 & 32.4 & 41.1 & 0.9 & 52.2 & 36.8 & 43.2 & 0.4 & 30.4 & 33.2 & 31.7 \\ 
compound & 0.4 & 74.2 & 66.2 & 69.9 & 0.6 & 72.5 & 63.6 & 67.8 & 4.4 & 84.7 & 51.6 & 64.1 \\ 
csubj & 0.2 & 77.0 & 52.5 & 62.4 & 0.3 & 88.1 & 57.3 & 69.4 & 0.2 & 45.9 & 31.3 & 37.2 \\ 
dep & 0.1 & 70.4 & 52.4 & 60.1 & 0.6 & 91.2 & 38.5 & 54.2 & 0.5 & 17.7 & 16.2 & 16.9 \\ 
discourse & 0.1 & 75.6 & 58.5 & 66.0 & 0.1 & 53.3 & 60.0 & 56.5 & 0.7 & 77.1 & 48.4 & 59.4 \\ 
foreign & 0.0 & 62.2 & 69.7 & 65.7 & 0.1 & 98.4 & 60.7 & 75.1 & 0.1 & 30.9 & 19.3 & 23.8 \\ 
goeswith & 0.0 & 35.7 & 29.4 & 32.3 & 0.1 & 75.0 & 19.6 & 31.1 & 0.0 & 26.1 & 16.7 & 20.3 \\ 
csubjpass & 0.0 & 100.0 & 73.9 & 85.0 & 0.0 & 93.3 & 71.2 & 80.8 & 0.1 & 87.5 & 19.7 & 32.2 \\ 
list & 0.0 & -- & -- & -- & 0.0 & 77.0 & 45.6 & 57.3 & 0.1 & 71.4 & 18.5 & 29.4 \\ 
remnant & 0.0 & 90.0 & 25.7 & 40.0 & 0.0 & 27.3 & 10.2 & 14.8 & 0.1 & 92.3 & 11.8 & 20.9 \\ 
reparandum & 0.0 & -- & -- & -- & 0.0 & -- & -- & -- & 0.1 & 100.0 & 34.6 & 51.4 \\ 
vocative & 0.0 & 55.6 & 31.3 & 40.0 & 0.0 & 57.4 & 52.9 & 55.1 & 0.1 & 84.5 & 58.6 & 69.2 \\ 
dislocated & 0.0 & 88.9 & 30.8 & 45.7 & 0.0 & 54.5 & 60.0 & 57.1 & 0.0 & 92.0 & 48.9 & 63.9 \\ \hline \hline
      \end{tabular}
     \caption{\footnotesize Precision, recall and f-score for different dependency labels for three groups of languages for the universal dependencies experiments in Table~\protect\ref{universal_results}: G1 (languages with $\text{UAS}\geq 80$), G2 (languages with $70\leq \text{UAS} <80$), G3 (languages with $\text{UAS}<70$). The rows are sorted by frequency in the G1 languages.}
    \label{tab:label_univ_accuracy}
\end{scriptsize}
\end{table*}

\section{Analysis} \label{analysis_section}

We conclude with some analysis of the accuracy of the method on different dependency types, 
across the different languages. Table~\ref{tab_en_dep_las} shows precision and
recall on different dependency types in English (using the Google
treebank). The improvements in accuracy when moving from the
delexicalized model to the Bible or Europarl model apply quite
uniformly across all dependency types, with all dependency labels
showing an improvement.

Table~\ref{tab:pos_dep_accuracy} shows the dependency accuracy sorted by part-of-speech
tag of the modifier in the dependency. We break the results into three
groups: G1 languages, where UAS is at least 80\% overall; G2
languages, where UAS is between 70\% and 80\%; and G3 languages, where
UAS is less than 70\%. There are some quite significant differences
in accuracy depending on the POS of the modifier word: for example in the G1
languages ADP, DET, ADJ, PRON and AUX all have over 85\% accuracy; in
contrast NOUN, VERB, PROPN, ADV all have accuracy that is less than
80\%. A very similar pattern is seen for the G2 languages, with ADP,
DET, ADJ, and AUX again having greater than 85\% accuracy, but 
NOUN, VERB, PROPN and ADV having lower accuracies. These results
suggest that difficulty varies quite signficantly depending on the
modifier POS, and different languages show the same patterns of
difficulty with respect to modifier POS.

Table ~\ref{tab:pos_head_accuracy} shows accuracy sorted by the POS tag of the {\em head} word of the dependency. By far the most frequent head POS
tags are NOUN, VERB, and PROPN (accounting for 85\% of all
dependencies). The table also shows that for all language groups G1,
G2, and G3, the f1 scores for NOUN, VERB and PROPN are generally
higher than the f1 scores for other head POS tags.

Finally, Table~\ref{tab:label_univ_accuracy} shows precision and recall for different
dependency labels for the G1, G2 and G3 languages. We again see quite
large differences in accuracy between different dependency labels. For
example in the G1 languages dependencies with the {\tt nmod} label,
the most frequent label, have 75.2\% f1 score; in contrast the second
most frequent label, {\tt case}, has 93.7\% f1 score. Other frequent labels
with low accuracy in the G1 languages are {\tt advmod}, {\tt conj},
and {\tt cc}.

\section{Related Work \label{related}}

There has recently been a great deal of work on syntactic transfer.
A number of methods \cite{zeman2008cross,mcdonald-petrov-hall:2011:EMNLP,cohen-das-smith:2011:EMNLP,naseem2012selective,tackstrom2013target,rosa-zabokrtsky:2015:ACL-IJCNLP} directly learn delexicalized models
that can be trained on universal treebank data from one or more
source languages, then applied to the target language. More recent
work has introduced cross-lingual representations---for example
cross-lingual word-embeddings---that can be used to improve
performance \cite{yuanregina15,guo-EtAl:2015:ACL-IJCNLP2,duong-EtAl:2015:CoNLL,duong-EtAl:2015:EMNLP,guo2016representation,ammar2016one}. These cross-lingual representations are 
usually learned from parallel translation data. We show results for
the methods of \cite{yuanregina15,guo2016representation,ammar2016one} in Table~\ref{tab_results2} of this paper.

The annotation projection approach, where dependencies from one
language are transferred through translation alignments to another
language, has been considered by several authors
\cite{hwa2005bootstrapping,ganchev-gillenwater-taskar:2009:ACLIJCNLP,mcdonald-petrov-hall:2011:EMNLP,ma-xia:2014:P14-1,rasooli-collins:2015:EMNLP,lacroix-EtAl:2016:N16-1,agic2016multilingual,schlichtkrull2017cross}. 
%The
%work of \newcite{rasooli-collins:2015:EMNLP} gives the best results of
%these methods on the Europarl data, as shown in
%Table~\ref{tab_results2}.

Other recent work \cite{tiedemann-agic-nivre:2014:W14-16,tiedemann2015improving,tiedemann2016synthetic} has considered treebank translation, where
a statistical machine translation system (e.g., MOSES \cite{koehn2007moses}) is
used to translate a source language treebank into the target language,
complete with reordering of the input sentence.
The lexicalization approach described in this paper is a simple
form of treebank translation, where we use a word-to-word translation
model. In spite of its simplicity, it is an effective approach.

A number of authors have considered incorporating universal syntactic
properties such as dependency order by selectively learning syntactic
attributes from similar source languages \cite{naseem2012selective,tackstrom2013target,yuanregina15,ammar2016one}. Selective sharing of
syntactic properties is complementary to our work; we used
a very limited form of selective sharing, through the WALS properties,
in our baseline approach. More recently, \newcite{wang2016galactic} have developed a synthetic treebank as a universal treebank to help learn parsers for new languages. \newcite{alonso2017parsing} try a very different approach in cross-lingual transfer by using a ranking approach.

A number of authors \cite{tackstrom2012cross,guo-EtAl:2015:ACL-IJCNLP2,guo2016representation} have introduced methods that learn
cross-lingual representations that are then used in syntactic transfer.
Most of these approaches introduce constraints to a clustering or
embedding algorithm that encourage words that are translations of each
other to have similar representations. Our method of deriving a
cross-lingual corpus (see Figure~\ref{cs_alg2}) is closely related to \cite{duong-EtAl:2015:CoNLL,gouws-sogaard:2015:NAACL-HLT,wick-et-al-2016}.%, who
%consider cross-lingual transfer of part-of-speech tags.

Our work has made use of dictionaries that are automatically extracted
from bilingual corpora. An alternative approach would be to use
hand-crafted translation lexicons, for example PanLex \cite{baldwin2010panlex}, which
covers 1253 language varieties, or Wiktionary (e.g., see \cite{durrett-pauls-klein:2012:EMNLP-CoNLL} for
an approach that uses Wiktionary for cross-lingual transfer). These
resources are a potentially very rich source of information; future
work should investigate whether they can give improvements in
performance.

\section{Conclusions}

We have described a method for cross-lingual syntactic transfer
that is effective in the scenario where a large amount of
translation data is not available. We have introduced a simple,
direct method for deriving cross-lingual clusters, and
for transferring lexical information across treebanks for
different languages. Experiments with the method show
that the method gives improved performance over previous work
that makes use of Europarl, a much larger translation corpus.
%Our method using the Europarl data improves over the state-of-the-art
%method of \newcite{rasooli-collins:2015:EMNLP} by 1.68\% in UAS and 2.14\% in LAS.

% and more sophisticated models for development of cross-lingual word representation, source treebank lexicalization and vocabulary expansion.

\begin{appendices}
\newcommand{\stp}[1]{\rho_{\sigma_{#1}}}
\newcommand{\bp}[1]{\rho_{\beta_{#1}}}
\newcommand{\sph}[1]{{\rho_{\sigma_{#1}^h}}}
\newcommand{\spl}[1]{{\rho_{\sigma_{#1}^l}}}
\newcommand{\spr}[1]{{\rho_{\sigma_{#1}^r}}}
\newcommand{\sprr}[1]{{\rho_{\sigma_{#1}^{r2}}}}
\newcommand{\spll}[1]{{\rho_{\sigma_{#1}^{l2}}}}
\newcommand{\bpll}[1]{{\rho_{\beta_{#1}^{l2}}}}
\newcommand{\sphh}[1]{{\rho_{{\sigma_{#1}^h}^h}}}
\newcommand{\bpl}[1]{{\rho_{\beta_{#1}^l}}}
\newcommand{\sw}[1]{\omega_{\sigma_{#1}}}
\newcommand{\bw}[1]{\omega_{\beta_{#1}}}
\newcommand{\swp}[1]{\omega\rho_{\sigma_{#1}}}
\newcommand{\bwp}[1]{\omega\rho_{\beta_{#1}}}
\newcommand{\svr}{{v_{\rightarrow}^\sigma}}
\newcommand{\svl}{{v_{\leftarrow}^\sigma}}
\newcommand{\bvl}{{v_{\leftarrow}^\beta}}
\newcommand{\swh}[1]{{\omega_{\sigma_{#1}^h}}}
\newcommand{\swhh}[1]{{\omega_{{\sigma_{#1}^h}^h}}}
\newcommand{\dis}{d}
\newcommand{\sdep}[1]{\iota_{\sigma_{#1}}}
\newcommand{\shdep}[1]{\iota_{\sigma_{#1}^h}}
\newcommand{\sldep}[1]{{\iota_{\sigma_{#1}^l}}}
\newcommand{\srdep}[1]{{\iota_{\sigma_{#1}^r}}}
\newcommand{\bldep}[1]{{\iota_{\beta_{#1}^l}}}
\newcommand{\slldep}[1]{{\iota_{\sigma_{#1}^{l2}}}}
\newcommand{\srrdep}[1]{{\iota_{\sigma_{#1}^{r2}}}}
\newcommand{\blldep}[1]{{\iota_{\beta_{#1}^{l2}}}}
\newcommand{\swl}[1]{{\omega_{\sigma_{#1}^l}}}
\newcommand{\swr}[1]{{\omega_{\sigma_{#1}^r}}}
\newcommand{\bwl}[1]{{\omega_{\beta_{#1}^l}}}
\newcommand{\swrr}[1]{{\omega_{\sigma_{#1}^{r2}}}}
\newcommand{\swll}[1]{{\omega_{\sigma_{#1}^{l2}}}}
\newcommand{\bwll}[1]{{\omega_{\beta_{#1}^{l2}}}}
\newcommand{\blab}[1]{{L_{\beta_{0}}^{#1}}}
\newcommand{\slab}[1]{{L_{\sigma_{0}}^{#1}}}

\newcommand{\stc}[1]{c^{#1}_{\sigma_{0}}}
\newcommand{\bc}[1]{c^{#1}_{\beta_{0}}}
\newcommand{\stx}[1]{x^{#1}_{\sigma_{0}}}
\newcommand{\bx}[1]{x^{#1}_{\beta_{0}}}

\section{Parsing Features}\label{apen_feat_sec} 
%In this appendix, we provide the full list of features that we use in this paper.  We only include cluster features, since others are exactly the same as the ones used in \cite[Table 1 and 2]{zhang-nivre:2011:ACL-HLT2011}.  The cluster features are obtained from the basic features \cite[Table 1]{zhang-nivre:2011:ACL-HLT2011} by replacing every POS of the first buffer and stack items with bit-strings of length 4 and 6 and every word of those items with the full bit-string. We use the following notations: $\sigma$: stack and $\beta$:  buffer (we use subscripts starting from zero to show to the index of the item in which the stack and buffer refers to); $\rho$: POS; $\omega$: word; $\omega\rho$: word-POS pair; $\iota$: dependency label; superscript $h$: head; superscript $l$: left-most modifier; superscript $l2$: second left-most modifier; superscript $r$: right-most modifier; superscript $r2$: second right-most modifier; $d$: distance between $\sigma_0$ and $\beta_0$; $c^{len}$ shows the word cluster feature first $len$ bit-string ($c^f$ shows the full bit-string). Also $x^{len}$ shows the cross-lingual word cluster features. 

We used	all features in~\cite[Table 1 and 2]{zhang-nivre:2011:ACL-HLT2011}, which
describes features based on the word and part-of-speech at various
positions on the stack and buffer of the transition system. In
addition, we expand the \newcite[Table 1]{zhang-nivre:2011:ACL-HLT2011} features to include
clusters, as follows: whenever a feature tests the part-of-speech for
a word in position 0 of the stack or buffer, we introduce features
that replace the part-of-speech	with the Brown clustering bit-string
of length 4 and	6. Whenever a feature tests for	the word identity
at position 0 of the stack or buffer, we introduce a cluster feature
that replaces the word with the	full cluster feature. We take the
cross product of all features corresponding to the choice of 4 or
6 length bit string for	part-of-speech features.

\end{appendices}
\bibliographystyle{acl2012} 
\bibliography{refs}

\end{document}